# Named Entity Recognition in Unstructured Medical Text Documents

Cole Pearson
*Dept. of Computer Science*
*University of Wisconsin – Eau Claire*
Eau Claire, United States
pearsocj7577@uwec.edu

Naeem Seliya
*Dept. of Computer Science*
*University of Wisconsin – Eau Claire*
Eau Claire, United States
seliyana@uwec.edu

Rushit Dave
*Dept. of Computer Science*
*University of Wisconsin – Eau Claire*
Eau Claire, United States
daver@uwec.edu

*Abstract*—Physicians provide expert opinion to legal courts on the medical state of patients, including determining if a patient is likely to have permanent or non-permanent injuries or ailments. An independent medical examination (IME) report summarizes a physician's medical opinion about a patient's health status based on the physician's expertise. IME reports contain private and sensitive information (Personally Identifiable Information or PII) that needs to be removed or randomly encoded before further research work can be conducted. In our study the IME is an orthopedic surgeon from a private practice in the United States. The goal of this research is to perform named entity recognition (NER) to identify and subsequently remove/encode PII information from IME reports prepared by the physician. We apply the NER toolkits of OpenNLP and spaCy, two freely available natural language processing platforms, and compare their precision, recall, and f-measure performance at identifying five categories of PII across trials of randomly selected IME reports using each model's common default parameters. We find that both platforms achieve high performance (f-measure > 0.9) at de-identification and that a spaCy model trained with a 70-30 train-test data split is most performant.

*Keywords—named entity recognition, de-identification, independent medical examination, natural language processing, machine learning.*

## I. INTRODUCTION

Independent medical examination (IME) reports are the summary documents prepared by independent examiners at the request of an insurance company or self-insured employer. These documents provide the medical professional's qualified opinion on the state of an injured worker's condition and assessment of any previous medical treatment the worker has undergone. Accordingly, IME reports contain a wide variety of sensitive information which must be removed or encoded to retain subject privacy before data-centric research may be performed on the documents.

Manually identifying and removing personally identifiable information (PII) from each report is inefficient and can be prohibitively time-consuming [1]. For this project, we were provided with a set of approximately 2000 IME reports from a medical practice in the United States. De-identifying these documents manually took on average between 15 and 30 minutes per report. To address this barrier to using large medical datasets, this project employs Named Entity Recognition (NER), a type of natural language processing. NER is a sequence tagging task which involves identifying and sorting tokens into predefined categories. Our models are trained using a selection of documents from our dataset in which PII has been manually tagged. The model is then applied to a secondary set of documents and its output compared to manually tagged versions of each report.

We used two open-source toolkits, OpenNLP [26] and spaCy [27], to train machine learning models capable of identifying and encoding PII in significantly less time required for manual processing. The results of the empirical case study presented in this paper indicate that a spaCy model trained with a 70-30 train-test data split achieves the highest recall (0.9239) and f-measure (0.9075) on average and is best suited to the task of IME report de-identification. An OpenNLP model with the same data split achieves slightly greater precision (0.9119) at the expense of recall and overall f-measure. We conclude that our findings demonstrate the feasibility of using freely available NER toolkits to automate the de-identification of unstructured medical text such as IME reports.

The remainder of this paper is structured as follows. Section II discusses related work in the general field of NER and more specifically NER on biomedical datasets. Section III discusses our methodology in the application of OpenNLP and spaCy. Section IV presents our findings, and Section V discusses the meaning of these results. Section VI considers the broader impact of our study and potential areas of future research.

## II. RELATED WORK

While we are not aware (to the best of our knowledge) of any other piece of research applying NER to de-identification of IME reports using the approach presented, there is substantial work in the general field of NER and in applying NER models to medical datasets. We take BioNER and de-identification as examples of the application of NER to medical data processing needs.

Alonso et al. [2] describe the current state of the field and presents many of the challenges common to NER tasks. The authors show that a parsing system with awareness of logical structures in a document can achieve higher performance. Various general and domain-specific NER systems are compared on appropriate datasets and shown to achieve f-measure scores ranging between 0.7103 and 0.9422 [2]. Another common practice in NER is the adaptation of models trained in high-resource domains to low-resource domains. This approach can allow researchers with a relatively small training sample to achieve better performance if a similar field with substantial annotated datasets exists [3]. Data augmentation is also useful when working with low-data source languages, such as Uyghur and Hungarian [1].

NER has been successfully applied to a wide variety of NER tasks. Recent published work shows high precision and recall using the Bi-LSTM architecture to assign informative word categories to road construction specifications [4]. A semi-supervised approach also showed moderate success parsing the education section of resumes [5].

In BioNER, the field aiming to identify biomedical entities such as medications and diseases, a variety of approaches have seen success in recent years. Leser and Hakenberg [6] survey the state of medical NER in 2005 and explains common challenges including rapidly changing domain knowledge and unstandardized abbreviations. More recently, many domain-specific and general-purpose models have overcome these obstacles, successfully identifying diseases, chemicals, anatomy, and other BioNER categories in unstructured reports [7][8][9]. Some success has been shown using regular expression-based systems, but results vary significantly by token category [10]. OpenNLP was shown to achieve moderate success (f-measure 0.715) at this task [11]. Ning and Bai [12] applied character-level as opposed to word-level vectors to solve issues resulting from low-frequency words.

Strategies have also been presented to improve performance in both large and limited datasets. Gao et al. [13] proposes transfer learning and semi-supervised training to aid where expert annotations are limited and achieves similar performance to that of a model trained on three to eight times the amount of data. Similarly, transfer learning and unsupervised training have been shown to improve f-measures on large, labelled datasets [14]. Other research seeks to improve the quality of existing datasets through label re-correction and knowledge distillation, iteratively modifying the NER model to yield trusted annotations [15]. This has also been done by developing stronger relationships between entities and context [16].

De-identification is the task of removing PII or other sensitive information from text to enable privacy-conscious use. Substantial research exists in using NER to de-identify unstructured medical notes and electronic health records (EHRs). Zuccon et al. [17] demonstrates that for highly regular token categories, NER solutions involving regular expressions can achieve good performance. Other work shows similar results using support vector machines to identify PII, noting that methods with local context perform better when presented with out-of-vocabulary target words and ambiguous terms [18]. A study using the standard 18 PII categories established by the Health Insurance Portability and Accountability Act in the United States achieved a high f-measure using the GATE text processing framework [19].

Seeking to improve NER performance and develop a gold corpus of medical discharge notes, one study found that iterative methods increased inter-annotator agreement, improving overall PII recognition performance [20]. Wellner et al. [21] concludes that an automated system supplemented by human review is the best current solution to the de-identification problem.

Medical NER has appeared as a precursor to derivative work, performed before data-focused analysis as in a study on radiological text reports [22]. Meystre et al. [23] considers what impact NER has on the resulting data contrasted with manual de-identification. The authors found minimal but not negligible loss of information, naming the resemblance of medication names with PII as a common cause for error.

### III. METHODOLOGY

The IME reports were provided to us by the medical practice in Microsoft Word (DOC) format. To develop a corpus of documents with tagged PII instances for use in training and evaluating our NER models, we manually opened and marked PII in a subset of our corpus as follows:

<START:categoryname>Token Content<END>

Here, "categoryname" represents the name of the classification group to which a PII instance belongs. "Token Content" represents the original text which comprises the instance.

*A. Named Entities*

PII instances were grouped into the following categories for annotation:

- Names: given names and surnames of individuals, including doctors and patients e.g., "John", "John Smith, M.D."

- Dates: date strings including at minimum a month and/or day, e.g., "March 1", "5/3/2000", "8/31"

- Places: names of organizations or locations, e.g., "Apple Town Medical Center", "Any Town, AL", "Alabama"

- Addresses: postal addresses, most commonly present in the header of a document, e.g., "123 Apple St., Any Town, AL 12345"

- Numbers: numerical identifiers such as license numbers and document numbers, e.g., "AL-12345", "123-456-7890"

In total, we annotated PII instances in 50 documents in the format shown above. The annotation process necessary to fully train and evaluate NER models provides an example of the need for an automated system. At an approximated average of 20 minutes per document, it took nearly 17 hours to complete this task. In comparison, a machine learning model takes a fraction of a second per document which is negligible for many small and medium-sized datasets and remains manageable at scales of hundreds or thousands of records.

These 50 documents were then divided randomly into two sets, one to be used in training the model and the other to be used in evaluating the trained model. We generated five training-evaluation sets each for the following splits, where the first number represents the proportion of documents which were in the training set and the second number represents the proportion in the evaluation set: 70-30, 66-34, 50-50.

For example, one of the five sets with a 70-30 split would have 70% × 50 = 35 reports in the set used to train the NER models and 50 − 35 = 15 in the set used for evaluation. Each train-test set randomized which documents were used in the training set and which in the test set.

### B. Microsoft Word Document Processing

The documents were provided in Microsoft Word 97-2003 document format (DOC files). Before the reports could be processed by OpenNLP or spaCy, we needed to extract the plain text content of the documents. For OpenNLP, a Java-based toolkit, this was accomplished using the HWPF module of the Apache POI library [24]. For spaCy, a Python-based toolkit, we used Textract to achieve a similar result [25].

### C. OpenNLP Pipeline

Once processed into unformatted text, the annotated training documents were passed directly into OpenNLP's sample loader, resulting in a NameSampleDataStream object. The PII annotation format described above is natively supported by OpenNLP, so the toolkit then parsed the documents to locate the manually annotated PII. To build the NER model, we used utilities provided in the TokenNameFinderFactory class and default the default parameters option. We called NameFinderME.train() directly, specifying a text language of English ("en") and passing our TokenNameFinderFactory and TrainingParameters Java objects as parameters. This results in a TokenNameFinderModel object, which is a trained model.

Next, we evaluated the model using OpenNLP's TokenNameFinderEvaluator. This requires reading in the annotated dataset via Apache POI then creating a NameFinderME Java object from the model and a TokenNameFinderEvaluator from the NameFinderME object. We could then call the evaluate() method of TokenNameFinderEvaluator followed by getFMeasure() to retrieve the precision, recall, and f-measure performance statistics [26]. We performed this testing process using annotated copies of both the training set and testing set as indicated in our results. This training and evaluation process was completed for each of the 15 scenarios representing five sets of random documents and three data splits.

### D. spaCy Pipeline

After being converted to plain text via Textract, we passed the annotated training documents into a custom-built engine to locate manual annotations, remove their marker text, and generate spaCy char_span objects. We set up the environment by creating an "nlp" variable: nlp=spacy.blank("en"). We then pass the unannotated text to nlp.make_doc to create a document representation and build a collection of PII instances with doc.char_span. We set the "ents" property of our document to the collection of PII instances. This is repeated for each document in the training set, and all documents are added to a DocBin object.

We performed the processing and evaluation of unannotated documents through spaCy's command line interface. This required us to export the DocBin representation to disk via its to_disk member function. We also write a configuration file which specifies that spaCy default parameters are to be used, generated using spaCy's online quickstart utility. We then invoked spaCy's train command with the DocBin export and config file, generating a trained model.

Model evaluation in spaCy was similarly performed via the command line interface. With the trained model file location and annotated dataset loaded as above, we invoked the evaluate command which outputs relevant statistics to a JSON file [27].

## IV. RESULTS

For each of the train-test data splits, we calculated the average precision, recall, and f-measure achieved by the spaCy and OpenNLP models. In Tables I, II, and III, "Train Data" refers to applying the model to identify PII in the same data that was used for training. "Test Data" refers to applying the model to those documents which were omitted from the training data.

When evaluated on the training data, all models tested had very high (>0.95) average f-measure, with spaCy scoring consistently higher than OpenNLP on each of the three measures taken. When comparing the models' performance on the test data set, differences correlating with the data split and model type become more apparent.

### A. Precision Performance Metric

Precision is defined as the fraction of PII instances found by the model which were consistent with the manually tagged copy, i.e.,

$$\text{Precision} = (A \cap B) / B \quad (1)$$

where $A$ is the set of PII instances present in the manually tagged version and $B$ is the set of PII instances identified by the machine learning model.

On the test data sets, the OpenNLP models consistently achieved higher average precision than the spaCy models. The data splits with more reports in the training set likewise performed better on average as seen in Table I. The model with highest average precision was the OpenNLP implementation trained and evaluated using a 70-30 split.

### B. Recall Performance Metric

Recall is defined as the fraction of PII instances present in the manually tagged copy which were successfully identified by the model, i.e.,

$$\text{Recall} = (A \cap B) / A \quad (2)$$

where $A$ is the set of PII instances present in the manually tagged version and $B$ is the set of PII instances identified by the machine learning model.

On the test data, spaCy achieved consistently higher average recall than OpenNLP at the data splits tested, and a preference again was shown for data splits with a higher training percentage as seen in Table II. The model with best average recall on the test data was spaCy with a 70-30 split.

*C. F-Measure Performance Metric*

F-measure is the harmonic mean of precision and recall, combining both into a single value. It is defined as

$$F\text{-measure} = 2 \times \text{precision} \times \text{recall} / (\text{precision} + \text{recall}) \quad (3)$$

On the test data, spaCy achieved higher average f-measure than OpenNLP at all data splits tested as seen in Table III. The spaCy model with a 70-30 data split showed the best overall performance at identifying PII.

As shown in Fig. 1, the most performant OpenNLP and spaCy models scored very similarly. While OpenNLP achieved higher precision, spaCy's higher recall led to a higher overall f-measure at our NER task.

TABLE I. SUMMARIZED AVERAGE MODEL PRECISION

| Data Split | OpenNLP Model | | spaCy Model | |
|---|---|---|---|---|
| | *Train Data* | *Test Data* | *Train Data* | *Test Data* |
| 70-30 | 0.9547 | 0.9119 | 0.9935 | 0.8961 |
| 66-34 | 0.9535 | 0.9053 | 0.9931 | 0.8904 |
| 50-50 | 0.9610 | 0.9050 | 0.9927 | 0.8834 |

TABLE II. SUMMARIZED AVERAGE MODEL RECALL

| Data Split | OpenNLP Model | | spaCy Model | |
|---|---|---|---|---|
| | *Train Data* | *Test Data* | *Train Data* | *Test Data* |
| 70-30 | 0.9579 | 0.8923 | 0.9962 | 0.9239 |
| 66-34 | 0.9588 | 0.8884 | 0.9968 | 0.9188 |
| 50-50 | 0.9606 | 0.8790 | 0.9967 | 0.9058 |

TABLE III. SUMMARIZED AVERAGE MODEL F-MEASURE

| Data Split | OpenNLP Model | | spaCy Model | |
|---|---|---|---|---|
| | *Train Data* | *Test Data* | *Train Data* | *Test Data* |
| 70-30 | 0.9562 | 0.9003 | 0.9948 | 0.9075 |
| 66-34 | 0.9560 | 0.8944 | 0.9950 | 0.9014 |
| 50-50 | 0.9607 | 0.8907 | 0.9947 | 0.8927 |

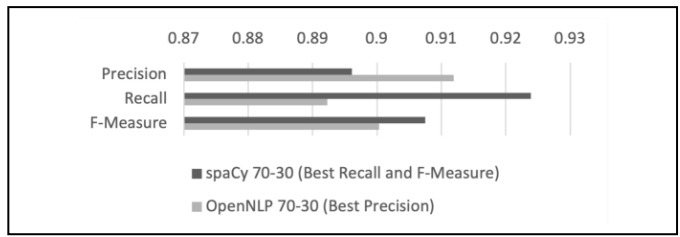

Fig. 1. Precision, Recall, and F-Measure of Highest Scoring Models in Each Evaluation Category

## V. DISCUSSION

Researchers seeking to use the content of IME reports as data often depend on manual de-identification methods which can be slow, expensive, and error prone. As researchers seek to include broader datasets and digitized health records make reports more widely available to the professional and academic community, the manual approach becomes infeasible and a barrier to continued work.

Our research demonstrates an NER system capable of identifying five general categories of PII for later encoding or removal with very high precision and recall. Using OpenNLP, an open-source natural language processing toolkit implemented in the Java programming language, we were able to achieve an average precision of 0.9119, average recall of 0.8923, and average f-measure of 0.9003 with a train-test split of 70-30 across 50 IME reports.

Approaching the same task with spaCy, an open-source toolkit written in Python, we saw a substantial improvement in recall at the expense of some precision. The most performant spaCy implementation, also using a 70-30 data split, achieved an average precision of 0.8961, average recall of 0.9239, and average f-measure of 0.9075.

Using f-measure as a common metric to average differences in precision and recall, the spaCy, 70-30 model appears as the best overall at identifying instances of PII in unstructured IME report text.

## VI. CONCLUSION

Using unstructured medical documents such as IME reports for data-centric research poses significant challenges to retain subject anonymity [28]. Although manual de-identification is possible, the time and resources necessary often make it a barrier to large-scale research [29]. NER via machine learning provides a highly performant alternative to manual de-identification which can be implemented with existing toolkits at comparatively low cost.

Our findings show OpenNLP and spaCy to be well-suited to this task despite their general intended purpose, with the best model (spaCy, 70-30) achieving an overall f-measure of 0.9075 across a sample of 50 IME reports from our dataset.

The process of training NER models based on pre-tagged sample data and applying the model to unfamiliar IME reports to locate PII takes a notably short amount of time compared to manual annotation of all documents [29]. In our work, a model could be trained from 25 to 35 documents in under one minute and applied to an unlabeled report in seconds. This low time requirement enabled the evaluation of numerous models for the purpose of comparing strategy. The machine learning de-identification techniques compared in this paper would be possible for researchers to run on standard personal computing hardware without the need for special compute resources.

A limitation of our findings is that the models in our analysis benefitted from common features across the dataset used, most notably that all IME reports were written by physicians associated with the same medical practice. The model fitting constructs could draw on similarities in document structure, such as use of the same headings and language patterns, to predict where PII instances would be likely to occur. Future work is necessary to determine whether this additional series of steps is in practice necessary to achieve performance similar to that presented in this paper.

Looking forward this approach would likely be sufficient for medium-sized tasks where researchers may benefit from a reduced manual de-identification workload. We have demonstrated that, for datasets which generally adhere to a similar writing format, de-identification through NER is possible with common toolkits. Processes like those compared in our study show that NER can provide a low-cost alternative to manual de-identification for researchers to proceed with data-centric work. Moreover, the NER process presented can be extended to non-medical documents, such as those related to IoT systems [30][31] and user authentication systems [32][33].

ACKNOWLEDGMENT

This work was supported in part by the Office of Research and Sponsored Programs (ORSP) and the Blugold Supercomputing Cluster at the University of Wisconsin – Eau Claire (UW-Eau Claire). The presented research findings and conclusions are solely those of the authors and are not representative of the ORSP or UW-Eau Claire.